\documentclass[conference]{IEEEtran}

\usepackage{amsmath}
\usepackage{amsfonts}
\usepackage{amssymb}
\usepackage{graphicx}
\usepackage{multirow}
\usepackage{wrapfig}
\usepackage{epsfig}
\usepackage{framed}
\usepackage{textcomp}
\usepackage{color}
\usepackage{booktabs}
\usepackage{pdflscape}
\usepackage[ruled]{algorithm2e}
\usepackage{url}

\definecolor{darkblue}{RGB}{47,68,117}
\definecolor{lightgray}{gray}{0.95}
\usepackage{subfigure}

\newcommand\blfootnote[1]{%
  \begingroup
  \renewcommand\thefootnote{}\footnote{#1}%
  \addtocounter{footnote}{-1}%
  \endgroup
}

\newcommand{\E}{\mathrm{E}}

\newcommand{\R}{\mathcal{R}}

\def\H{{\mathbf H}}

\newcommand{\popsize}{n}

\newcommand{\bmx}[0]{\begin{bmatrix}}
\newcommand{\emx}[0]{\end{bmatrix}}

\newcommand{\vects}[1]{\boldsymbol{#1}}
\newcommand{\vect}[1]{\mathbf{#1}}
\newcommand{\matr}[1]{\mathbf{#1}}
\newcommand{\var}[0]{\operatorname{Var}}
\newcommand{\cov}[0]{\operatorname{Cov}}
\newcommand{\diag}[0]{\operatorname{diag}}

\newcommand{\vb}[0]{\vect{b}}
\newcommand{\vd}[0]{\vect{d}}

\newcommand{\vh}[0]{\vect{h}}

\newcommand{\vt}[0]{\vect{t}}
\newcommand{\vg}[0]{\vect{g}}

\newcommand{\mW}[0]{\matr{W}}

\newcommand{\mH}[0]{\matr{H}}

\newcommand{\mA}{\matr{A}}

\newcommand{\TT}[0]{{\vects{\theta}}}

\newcommand{\f}[0]{\text{f}}
\newcommand{\g}[0]{\text{g}}
\DeclareMathOperator*{\argmin}{arg\,min}

\newcommand{\be}{\begin{equation}}
\newcommand{\ee}{\end{equation}}
\newcommand{\bea}{\begin{eqnarray}}
\newcommand{\eea}{\end{eqnarray}}
\newcommand{\eqnarr}{\begin{eqnarray}}
\newcommand{\eqnend}{\end{eqnarray}}

\DeclareMathSymbol{\umu}{\mathalpha}{operators}{0}

\begin{document}
%
\title{A Robust Adaptive Stochastic \\ Gradient Method for Deep Learning}

\author{\IEEEauthorblockN{Caglar Gulcehre*}
\IEEEauthorblockA{Universit\'e de Montr\'eal}
\and
\IEEEauthorblockN{Jose Sotelo*}
\IEEEauthorblockA{Universit\'e de Montr\'eal}
\and
\IEEEauthorblockN{Marcin Moczulski}
\IEEEauthorblockA{University of Oxford}
\and
\IEEEauthorblockN{Yoshua Bengio}
\IEEEauthorblockA{Universit\'e de Montr\'eal}}

\maketitle
\blfootnote{* denotes equal contribution.}

\begin{abstract}
    Stochastic gradient algorithms are the main focus of large-scale optimization problems and led to
    important successes in the recent advancement of the deep learning algorithms. The convergence
    of SGD depends on the careful choice of learning rate and the amount of the noise in
    stochastic estimates of the gradients. In this paper, we propose an adaptive learning rate
    algorithm, which utilizes stochastic curvature information of the loss function for
    automatically tuning the learning rates. The information about the element-wise curvature of
    the loss function is estimated from the local statistics of the stochastic first order
    gradients. We further propose a new variance reduction technique to speed up the convergence.
    In our experiments with deep neural networks, we obtained better performance compared to the
    popular stochastic gradient algorithms. \footnote{This paper is an extension/update of our
    previous paper \cite{gulcehre2014adasecant}.}
\end{abstract}


\IEEEpeerreviewmaketitle

\section{Introduction}

We develop an automatic stochastic gradient algorithm which reduces the burden of extensive
hyper-parameter search for the optimizer. Our proposed algorithm exploits a lower variance
estimator of curvature of the cost function and uses it to obtain an automatically tuned adaptive
learning rate for each parameter.

In deep learning and numerical optimization literature, several papers suggest using a diagonal
approximation of the Hessian (second derivative matrix of the cost function with respect to
parameters), in order to estimate optimal learning rates for stochastic gradient descent over high
dimensional parameter spaces \cite{becker1988improving,schaul2012no,lecun1993automatic}.  A
fundamental advantage of using such approximation is that inverting such approximation can be a
trivial and cheap operation.  However generally, for neural networks, the inverse of the diagonal Hessian is usually
a bad approximation of the diagonal of the inverse of Hessian.  For example, obtaining a diagonal
approximation of Hessian are the Gauss-Newton matrix \cite{lecun2012efficient} or by finite
differences \cite{schaul2013adaptive}. Such estimations may however be very sensitive to the noise
coming from the Monte-Carlo estimates of the gradients.  \cite{schaul2012no} suggested a reliable
way to estimate the local curvature in the stochastic setting by keeping track of the variance and
average of the gradients.

We propose a different approach: instead of using a diagonal estimate of Hessian, to estimate
curvature along the direction of the gradient and we apply a new variance reduction technique to
compute it reliably. By using root mean square statistics, the variance of gradients are reduced
adaptively with a simple transformation. We keep track of the estimation of curvature using a
technique similar to that proposed by \cite{schaul2012no}, which uses the variability of the
expected loss. Standard adaptive learning rate algorithms only scale the gradients, but regular
Newton-like second order methods, can perform more complicate transformations, e.g. rotating the
gradient vector. Newton and quasi-newton methods can also be invariant to affine transformations
in the parameter space. \textbf{AdaSecant} algorithm is basically a stochastic rank-$1$
quasi-Newton method. But in comparison with other adaptive learning algorithms, instead of just
scaling the gradient of each parameter, AdaSecant can also perform an affine transformation on
them.

\section{Directional Secant Approximation}
\label{sec:dir_sec_approx}

Directional Newton is a method proposed for solving equations with multiple variables\cite{levin2002directional}.  The advantage of directional Newton method proposed in\cite{levin2002directional}, compared to Newton's method is that, it does not require a matrix inversion and still maintains a quadratic rate of convergence.

In this paper, we develop a second-order directional Newton method for nonlinear optimization. Step-size $\vt^k$ of update $\Delta^k$ for step $k$ can be written as if it was a diagonal matrix:

\begin{align}
\label{eqn:dir_newton}
\Delta^k  & =  - \vt^k \odot \nabla_{\TT} \f(\TT^k), \\
& = - \diag(\vt^k) \nabla_{\TT} \f(\TT^k), \\
& = - \diag(\vd^k)(\diag(\H \vd^k))^{-1} \nabla_{\TT} \f(\TT^k).
\end{align}
where $\TT^k$ is the parameter vector at update $k$, $\f$ is the objective function and $\vd^k$ is a
unit vector of direction that the optimization algorithm should follow.
Denoting by $\vh_i=\nabla_{\TT}\frac{\partial \f(\TT^k)}{\partial \theta_i}$ the $i^{th}$ row of the Hessian matrix $\H$
and by $\nabla_{\TT_i} f(\TT^k)$ the $i^{th}$ element of the gradient vector at update $k$,
a reformulation of Equation \ref{eqn:dir_newton} for each diagonal element of the step-size $\diag(\vt^k)$ is:
\begin{align}
  \Delta_i^k & = - t_i^k \nabla_{\TT_i }\f(\TT^k),  \\
  & = - d_i^k \frac{\nabla_{\TT_i }\f(\TT^k)}{\vh^k_i\vd^k}.
\end{align}
so effectively
\begin{equation}
  t_i^k=\frac{d_i^k}{\vh^k_i\vd^k}.
\end{equation}

We can approximate the per-parameter learning rate $t_{i}^k$ following \cite{an2005directional} using finite differences:
\begin{align}
\label{eqn:step_size_eq}
t_i^k &= \frac{d_i^k}{\vh_i^k \vd^k}, \\
      &= \lim_{|\Delta_{i}^k| \rightarrow 0} \frac{\Delta_{i}^k}{\nabla_{\TT_i} \f(\TT^k + \Delta^k)
      - \nabla_{\TT_i} \f(\TT^k)}, \text{for every } i . 
\end{align}

Let us note that alternatively one might use the R-op to compute the Hessian-vector product for the denominator in Equation \ref{eqn:step_size_eq} \cite{schraudolph2002fast}.

To choose a good direction $\vd^k$ in the stochastic setting, we use block-normalized gradient vector that the parameters of each layer is considered as a block and for each weight matrix $\mW^i_k$ and bias vector $\vb^i_k$ for $\TT=\left\{\mW^i_k, \vb^i_k\right\}_{i=1\cdots k}$ at each layer $i$ and update $k$,
∏$\vd_k = \left[ \vd_{\mW^0_k}^k \vd_{\vb^0_k}^k \cdots \vd_{\vb^l_k}^k\right]$ for a neural
network with $l$ layers.

The update step is defined as $\Delta_{i}^k = t_i^k d_i^k$. The per-parameter learning rate
$t_i^k$ can be estimated with the finite difference approximation,

\begin{equation}
t_i^k \approx \frac{\Delta_{i}^k}{\nabla_{\TT_i} \f(\TT^k + \Delta^k) - \nabla_{\TT_i} \f(\TT^k)},
\end{equation}
since, in the vicinity of the quadratic local minima,
\begin{align}
    \nabla_{\TT} \f(\TT^k + \Delta^k) - \nabla_{\TT} \f(\TT^k) \approx \H^{k} \Delta^k ,
\end{align}
We can therefore recover $\vt^k$ as
\begin{equation}
\vt^k = \diag(\Delta^k)(\diag(\H^{k} \Delta^k))^{-1}.
\end{equation}
The directional secant method basically scales the gradient of each parameter with the curvature along the direction of the gradient vector
and it is numerically stable.

\section{Relationship to the Diagonal Approximation to the Hessian}
Our secant approximation of the gradients are also very closely tied to diagonal approximation of
the Hessian matrix. Considering that $i^{th}$ diagonal entry of the Hessian matrix can be denoted
as, $\mH_{ii} = \frac{\partial^2 \f(\TT)}{\partial \TT_i^2}$. By using the finite differences, it
is possible to approximate this with as in Equation \ref{eqn:diag_hessian_finite},ıı

\begin{equation}
    \label{eqn:diag_hessian_finite}
    \mH_{ii} = \lim_{|\Delta| \rightarrow 0} \frac{\nabla_{\TT_i} \f(\TT + \Delta) -
    \nabla_{\TT_i}\f(\TT)}{\Delta_i},
\end{equation}

Assuming that the diagonal of the Hessian is denoted with $\mA$ matrix, we can see the equivalence:
\begin{equation}
    \label{eqn:diag_approx_hessian}
    \mA~\approx~ \diag(\nabla_{\TT} \f(\TT + \Delta) - \nabla_{\TT} \f(\TT))\diag(\Delta)^{-1}.
\end{equation}
The Equation \ref{eqn:diag_approx_hessian} can be easily computed in a stochastic setting from the
consecutive minibatches.

\section{Variance Reduction for Robust Stochastic Gradient Descent}
\label{sec:var_reduction}
Variance reduction techniques for stochastic gradient estimators have been well-studied
in the machine learning literature. Both \cite{wang2013variance} and \cite{johnson2013accelerating} proposed new
ways of dealing with this problem. In this paper, we proposed a new variance reduction technique
for stochastic gradient descent that relies only on basic statistics related to the gradient.
Let $\g_i$ refer to the $i^{th}$ element of the gradient vector $\vg$  with respect to the parameters
$\TT$ and $\E[\cdot]$ be an expectation taken over minibatches and different trajectories of
parameters.

We propose to apply the following transformation to reduce the variance of the stochastic
gradients:
\begin{equation}
\label{eqn:tilde_gamma_i_def}
\tilde{g_i}=\frac{g_i + \gamma_i\E[g_i]}{1+\gamma_i},
\end{equation}
%
%
where $\gamma_i$ is strictly a positive real number.
Let us note that:
\begin{align}
\E[\tilde{g_i}] = \E[g_i]
 \text{  and }
 \var(\tilde{g_i}) = \frac{1}{(1+\gamma_i)^2}\var(g_i).
\end{align}

The variance is reduced by a factor of $(1+\gamma_i)^2$ compared to $\var(g_i)$.

In practice we do not have access to $\E[g_i]$, therefore
a biased estimator $\overline{g_i}$ based on past values of $g_i$ will be used instead.
We can rewrite the $\tilde{g_i}$ as:
\begin{equation}
\tilde{g_i} = \frac{1}{1+\gamma_i}g_i + (1 - \frac{1}{1+\gamma_i}) \E[g_i],
\end{equation}
After substitution $\beta_i = \frac{1}{1+\gamma_i}$, we will have:
\begin{equation}
 \tilde{g_i} = \beta_i g_i + (1 - \beta_i) \E[g_i].
\end{equation}
By adapting $\gamma_i$ or $\beta_i$, it is possible to control the influence of
high variance, unbiased $g_i$ and low variance, biased $\overline{g_i}$ on $\tilde{g_i}$.
Denoting by $\vg^{\prime}$ the stochastic gradient obtained on the next minibatch,
the $\gamma_i$ that well balances those two influences is the one that keeps the $\tilde{g_i}$
as close as possible to the true gradient $\E[g_i^{\prime}]$ with $g_i^{\prime}$ being the only sample of
$\E[g_i^{\prime}]$ available. We try to find a regularized $\beta_i$, in order to obtain a
smoother estimate of it and this yields us to more more stable estimates of $\beta_i$. $\lambda$
is the regularization coefficient for $\beta$.

%
\begin{equation}
\label{eqn:exp_gamma_cri}
\argmin_{\beta_i} \E[||\tilde{g_i} - g_i^{\prime}||^2_2] + \lambda (\beta_i)^2.
\end{equation}
It can be shown that this a convex problem in $\beta_i$ with a closed-form solution (details in
appendix) and we can obtain the $\gamma_i$ from it:

\begin{equation}
\label{eqn:gamma_i_formula}
\gamma_i = \frac{\E[(g_i - g_i^{\prime})(g_i - \E[g_i])]}{\E[(g_i-\E[g_i])(g^{_i\prime}-\E[g_i]))] + \lambda},
\end{equation}
As a result, to estimate $\gamma$ for each dimension, we keep track of a estimation of
$\frac{\E[(g_i - g_i^{\prime})(g_i - \E[g_i])]}{\E[(g_i - \E[g_i])(g_i^{\prime}-\E[g_i]))] + \lambda}$
during training. The necessary and sufficient condition here, for the variance reduction is to
keep $\gamma$ positive, to achieve a positive estimate of $\gamma$ we used the root mean square
statistics for the expectations.

\section{Blockwise Gradient Normalization}

It is very well-known that the repeated application of the non-linearities can cause the gradients
to vanish \cite{bengio1994learning,hochreiter2001gradient}. Thus, in order to tackle this
problem, we normalize the gradients coming into each block/layer to have norm 1.
Assuming the normalized gradient can be denoted with $\tilde{\vg}$, it can be computed as,
$\tilde{\vg} = \frac{\vg}{||\E[\vg]||_2}$. We estimate, $\E[\vg]$ via moving averages.

Blockwise gradient normalization of the gradient adds noise to the gradients, but in practice we did
not observe any negative impact of it.  We conjecture that this is due to the angle between the
stochastic gradient and the block-normalized gradient still being less than $90$ degrees.

\section{Adaptive Step-size in Stochastic Case}

In the stochastic gradient case, the step-size of the directional secant can be computed by using
an expectation over the minibatches:
\begin{equation}
\label{eqn:adapt_stepsize}
\E_k[t_{i}] = \E_k[\frac{\Delta_i^k}{\nabla_{\TT_i} \f(\TT^k +  \Delta^k) - \nabla_{\TT_i}
\f(\TT^k)}].
\end{equation}
The $E_k[\cdot]$ that is used to compute the secant update, is taken over the minibatches at the past
values of the parameters.

Computing the expectation in Equation\ref{eqn:adapt_stepsize} was numerically unstable in stochastic
setting. We decided to use a more stable second order Taylor approximation of
Equation \ref{eqn:adapt_stepsize} around $(\sqrt{\E_k[(\alpha_i^k)^2]}, \sqrt{\E_k[(\Delta_i^k)^2]})$, with $\alpha_i^k = \nabla_{\TT_i} \f(\TT^k + \Delta^k) - \nabla_{\TT_i} \f(\TT^k)$.
Assuming  $\sqrt{\E_k[(\alpha_i^k)^2]} \approx \E_k[\alpha_i^k]$ and $\sqrt{\E_k[(\Delta_i^k)^2]} \approx \E_k[\Delta_i^k]$ we obtain always non-negative approximation of $\E_k[t_{i}]$:
\begin{align}
\label{eqn:step_size_stoc_der1}
& \E_k[t_{i}] \approx \frac{\sqrt{\E_k[(\Delta_i^k)^2]}}{\sqrt{\E_k[(\alpha_i^k)^2]}} -
\frac{\cov(\alpha_i^k,\Delta_i^k)}{\E_k[(\alpha_i^k)^2]}.
\end{align}
In our experiments, we used a simpler approximation,
which in practice worked as well as formulations in Equation\ref{eqn:step_size_stoc_der1}:
\begin{equation}
\label{eqn:step_size_stoc2}
\E_k[t_{i}] \approx \frac{\sqrt{\E_k[(\Delta_i^k)^2]}}{\sqrt{\E_k[(\alpha_i^k)^2]}} -
\frac{\E_k[\alpha_i^k\Delta_i^k]}{\E_k[(\alpha_i^k)^2]}.
\end{equation}

\section{Algorithmic Details}
\subsection{Approximate Variability}
To compute the moving averages as also adopted by \cite{schaul2012no},
we used an algorithm to dynamically decide the time constant based on the step size being taken.
As a result algorithm that we used will give bigger weights to the updates that have large step-size
and smaller weights to the updates that have smaller step-size.

By assuming that $\bar{\Delta}_i[k]\approx\E[\Delta_i]_k$,
the moving average update rule for $\bar{\Delta}_i[k]$ can be written as,
\begin{align}
    \label{eqn:approx_var}
    & \bar{\Delta}_i^2[k] = (1~-~\tau_i^{-1}[k])\bar{\Delta}_i^2[k-1] + \tau_i^{-1}[k](t_i^k
    \tilde{\vg}_i^k),
\end{align}
and,
\begin{align}
    \bar{\Delta}_i[k] = \sqrt{\bar{\Delta}_i^2[k]}.
\end{align}
This rule for each update assigns a different weight to each element of the gradient vector .
At each iteration a scalar multiplication with $\tau_i^{-1}$ is performed and $\tau_i$ is adapted using the following equation:
\begin{equation}
    \label{eqn:approx_var_update}
    \tau_i[k] = (1~-~\frac{\E[\Delta_i]_{k-1}^2}{\E[(\Delta_i)^2]_{k-1}})\tau_i[k-1] + 1~.
\end{equation}
%
\subsection{Outlier Gradient Detection}
%
Our algorithm is very similar to \cite{schaul2013adaptive}, but instead of
incrementing $\tau_i[t+1]$ when an outlier is detected, the time-constant is reset
to $2.2$. Note that when $\tau_i[t+1] \approx 2$, this assigns approximately the same amount of
weight to the current and the average of previous observations.  This mechanism made learning more stable,
because without it outlier gradients saturate $\tau_i$ to a large value.
\subsection{Variance Reduction}
The correction parameters $\gamma_i$ (Equation\ref{eqn:gamma_i_formula}) allows for a fine-grained variance reduction for each parameter independently.
%
The noise in the stochastic gradient methods can have advantages both in terms of generalization and optimization.
It introduces an exploration and exploitation trade-off, which can be controlled
by upper bounding the values of $\gamma_i$ with a value $\rho_i$, so that
thresholded $\gamma_i^{\prime} = \min(\rho_i, \gamma_i)$.
%

We block-wise normalized the gradients of each weight matrix and bias vectors in $\vg$ to compute the
$\tilde{\vg}$ as described in Section~\ref{sec:dir_sec_approx}.
That makes AdaSecant scale-invariant, thus more robust to the scale of the inputs
and the number of the layers of the network. We observed empirically that it was easier to train very deep neural networks with block
normalized gradient descent. In our experiments, we fixed $\lambda$ to $1e-5$.
\section{Improving Convergence}
Classical convergence results for SGD are based on the conditions:
\begin{equation}
\sum_i (\eta^{(i)})^2 < \infty \text{ and }   \sum_i \eta^{(i)} = \infty
\end{equation}
such that the learning rate $\eta^{(i)}$ should decrease \cite{robbins1951stochastic}. Due to the noise in the estimation of
adaptive step-sizes for AdaSecant, the convergence would not be guaranteed. To ensure it,
we developed a new variant of Adagrad \cite{duchi2011adaptive} with thresholding,
such that each scaling factor
is lower bounded by $1$. Assuming  $a_i^k$ is the accumulated norm of all past gradients for $i^{th}$ parameter at update $k$, it is thresholded from below ensuring that the algorithm will converge:
\begin{equation}
  a_i^k = \sqrt{\sum_{j=0}^k (g_i^j)^2},
\end{equation}
and
\begin{equation}
  \rho_i^k = \text{maximum}(1, a_i^k),
\end{equation}
giving
\begin{equation}
  \Delta_i^k = \frac{1}{\rho_i}\eta_i^k\tilde{\vg}_i^k.
\end{equation}
In the initial stages of training, accumulated norm of the per-parameter gradients can be less than $1$. If the accumulated per-parameter norm of a gradient is less than $1$, Adagrad will augment
the learning-rate determined by AdaSecant for that update, i.e.
$\frac{\eta_i^k}{\rho_i^k} > \eta_i^k$ where $\eta_i^k=\E_k[t_i^k]$ is the per-parameter learning rate determined
by AdaSecant. This behavior tends to create unstabilities during the training
with AdaSecant. Our modification of the Adagrad algorithm is to ensure that, it will
 reduce the learning rate determined by the AdaSecant algorithm at each update, i.e. $\frac{\eta_i^k}{\rho_i^k} \le
 \eta_i^k$ and the learning rate will be bounded. At the beginning of the training, parameters of a
 neural network can get $0$-valued gradients, e.g. in the existence of dropout and ReLU units.
 However this phenomena can cause the per-parameter learning rate scaled by Adagrad to be
 unbounded.

In Algorithm \ref{alg:adasecant}, we provide a simple pseudo-code of the AdaSecant algorithm.
\begin{algorithm}[htb]
\SetKwInOut{Input}{input}
\SetKwInOut{Output}{output}
\label{alg:adasecant}
\caption{AdaSecant: minibatch-AdaSecant for adaptive learning rates with variance reduction}

 \Repeat{stopping criterion is met}{
  draw $\popsize$ samples,
  compute the gradients $\vg^{(j)}$ where $\vg^{(j)} \in \R^{n}$ for each minibatch $j$,
  $\vg^{(j)}$ is computed as,~$\frac{1}{\popsize}\sum_{k=1}^{\popsize}\nabla_{\TT}^{(k)}\f(\TT)$ \\
  estimate $\E[\vg]$ via moving averages. \\
  block-wise normalize gradients of each weight matrix and bias vector \\
  \For{\text{parameter} $i \in \{1, \ldots, n\}$}{
    compute the correction term by using, $\gamma^k_i =\frac{\E[(g_i - g_i^{\prime})(g_i - \E[g_i])]_k}{\E[(g_i-\E[g_i])(g_i^{\prime}-\E[g_i]))]_k}$ \\
    compute corrected gradients $\tilde{g_i}=\frac{g_i+\mathbf{\gamma_i}\E[g_i]}{1+\mathbf{\gamma_i}}$ \\
    \vspace{0.5em}
    \If{$|g_i^{(j)}- \E[g_i]| > 2 \sqrt{\E[(g_i)^2]-(\E[g_i])^2} \;\;\; \operatorname{or} \;\;\;
        \left|\alpha_i^{(j)} - \E[\alpha_i]\right| > 2 \sqrt{\E[(\alpha_i)^2] - (\E[\alpha_i])^2}$}
    {\vspace{0.5em}
    reset the memory size for outliers $\tau_i \leftarrow 2.2$}{}
    \vspace{0.5em}
    update moving averages according to Equation \ref{eqn:approx_var} \\
\vspace{0.5em}
  estimate learning rate $\;\;\displaystyle\eta_i^{(j)} \leftarrow
      \frac{\sqrt{\E_k[(\Delta_i^{(k)})^2]}}{\sqrt{\E_k[(\alpha_i^k)^2]}} -
      \frac{\E_k[\alpha_i^k\Delta_i^k]}{\E_k[(\alpha_i^k)^2]}$\\
  \vspace{0.5em}
  update memory size as in Equation \ref{eqn:approx_var_update}\\
 \vspace{0.5em}
 update parameter $\;\; \theta_i^j \leftarrow \theta_i^{j-1} - \eta_i^{(j)}\cdot \tilde{g}_i^{(j)}$\\
 }
 }

\end{algorithm}
\section{Experiments}
We have run experiments on character-level PTB with GRU units, on MNIST with Maxout Networks
\cite{goodfellow2013maxout} and on handwriting synthesis using the IAM-OnDB dataset \cite{LiwickiB05a}. We
compare AdaSecant with popular stochastic gradient learning algorithms: Adagrad, RMSProp
\cite{graves2013generating}, Adadelta \cite{zeiler2012adadelta}, Adam \cite{Kingma2014} and
SGD+momentum (with linearly decaying learning rate). AdaSecant performs as well or better as
carefully tuned algorithms for all these different tasks.

\subsection{Ablation Study}

\begin{table}[htp]
\centering

 \begin{tabular}{|c c c|}
 \hline
 Model & Train Log-Loss & Valid Log-Loss \\ [0.5ex]
 \hline\hline
 Adam with 3e-4 learning rate & -1.827 & -1.743 \\
 \hline
 Adam with 1e-4 learning rate & -1.780 & -1.713 \\
  \hline
 Adam with 5e-4 learning rate & -1.892 & -1.773 \\
 \hline
 AdaSecant  & \textbf{-1.881} & -1.744 \\
 \hline
 AdaSecant, no VR & -1.876 & -1.743 \\
  \hline
 AdaSecant, no AG & -1.867 & -1.738 \\
  \hline
 AdaSecant, no BN & -1.857 & -1.784 \\
  \hline
 AdaSecant, no OD & -1.780 & -1.726 \\
  \hline
 AdaSecant, no VR, no AG & -1.848 & -1.744 \\
  \hline
 AdaSecant, no VR, no BN & -1.844 & -1.777 \\
  \hline
 AdaSecant, no VR, no OD & -1.479 & -1.442 \\
  \hline
  AdaSecant, no AG, no BN & -1.878 & \textbf{-1.786} \\
  \hline
 AdaSecant, no AG, no OD & -1.723 & -1.674 \\
  \hline
 AdaSecant, no BN, no OD & -1.814 & -1.764 \\
  \hline
 AdaSecant, no AG, no BN, no OD & -1.611 & -1.573 \\
  \hline
 AdaSecant, no VR, no BN, no OD & -1.531 & -1.491 \\
   \hline
 AdaSecant, no VR, no AG, no OD & unstable & unstable \\
   \hline
 AdaSecant, no VR, no AG, no BN & -1.862 & 1.75 \\
 \hline
\end{tabular}
\vspace{1mm}
 \caption{Summary of results for the handwriting experiment.
 We report the best validation log-loss that we found for each model using early stopping. We also report the corresponding train log-loss. In all cases, the log-loss is computed per data point.
 \label{tab:results}
 }
\end{table}

In this section, we decompose the different parts of the algorithm to measure the effect they have
in the performance. For this comparison, we trained a model to learn handwriting synthesis on
IAM-OnDB dataset. Our model follows closely the architecture introduced in \cite{graves2013generating} with two
modifications. First, we use one recurrent layer of size 400 instead of three. Second, we use GRU
\cite{cho2014learning} units instead of LSTM \cite{Hochreiter1997} units. Also, we use a different
symbol for each of the 87 different characters in the dataset. The code for this experiment is
available online.\footnote{\url{https://github.com/sotelo/scribe}}

\begin{figure}[htp]
\centering
\includegraphics[width=8cm]{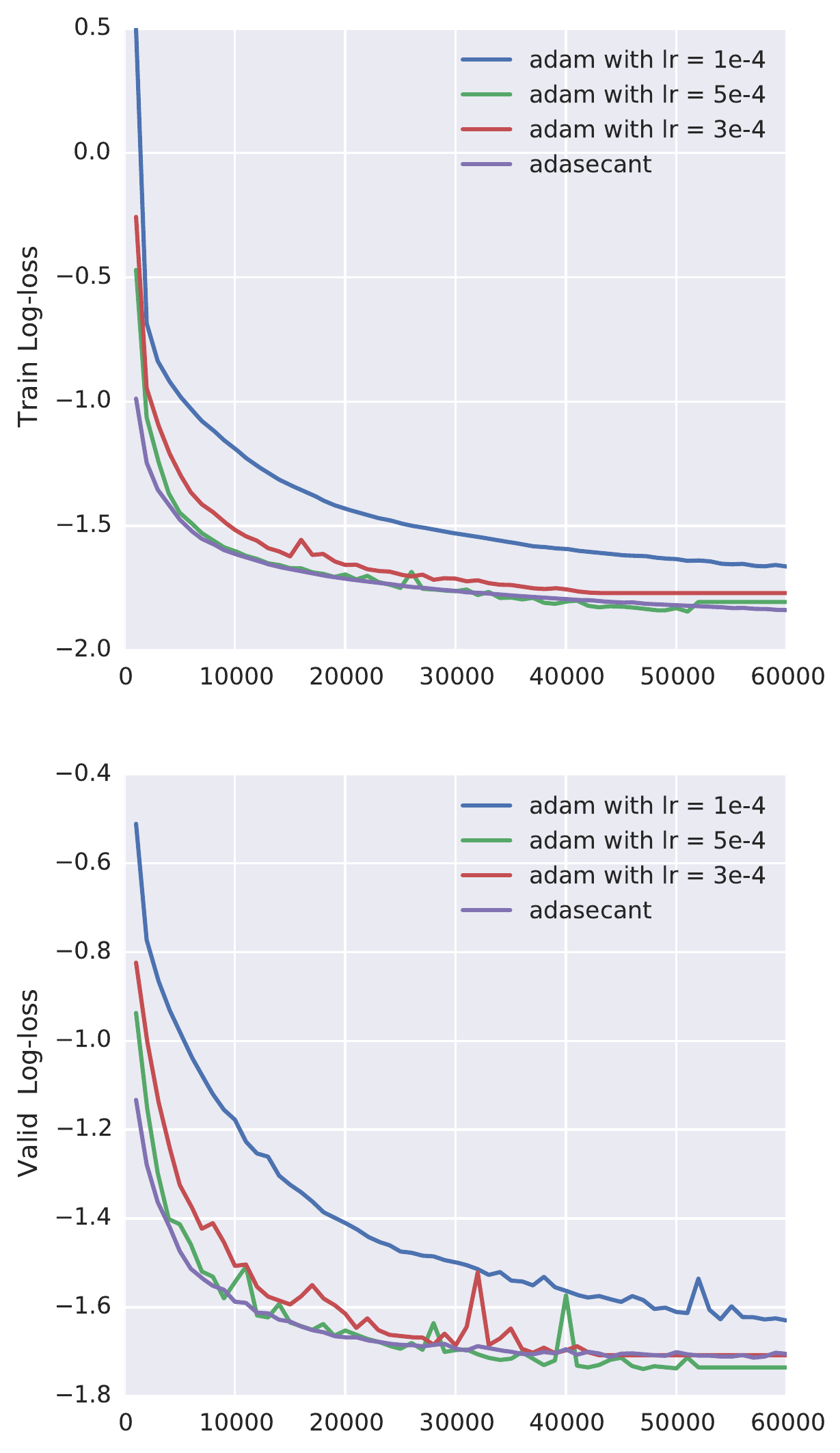}
\caption{Baseline comparison against Adam. AdaSecant performs as well as Adam with a carefully tuned learning rate.}
\label{fig:adam_vs_adasecant}
\end{figure}

We tested different configurations that included taking away the use of Variance Reduction (VR),
Adagrad (AG), Block Normalization (BN), and Outlier Detection (OD). Also, we compared against ADAM
\cite{Kingma2014} with different learning rates in Figure~\ref{fig:adam_vs_adasecant}. There, we
observe that adasecant performs as well as Adam with a carefully tuned learning rate.

\begin{figure}[htp]
\centering
\includegraphics[width=8cm]{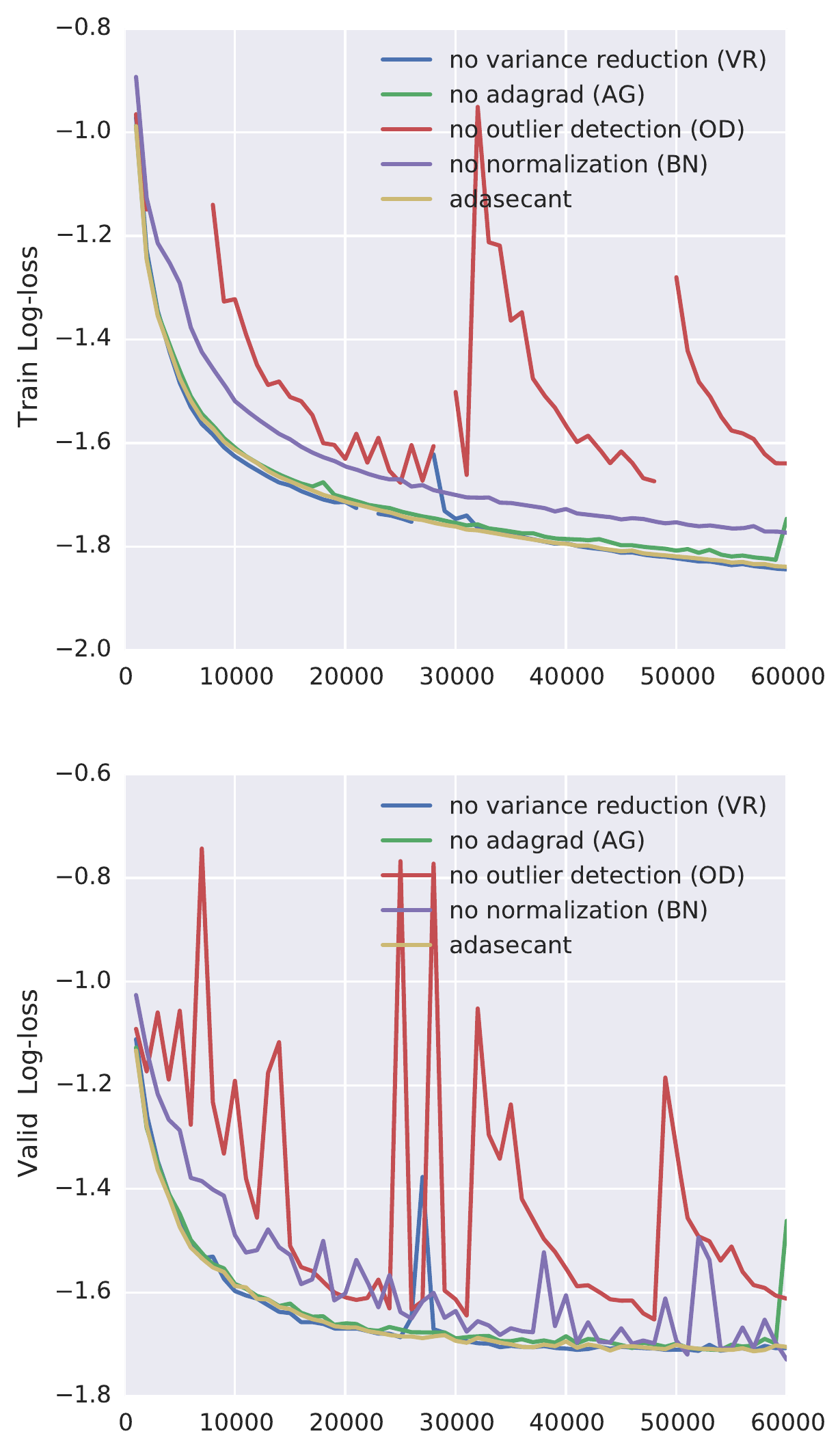}
\caption{Deactivating one component at a time. BN provides a small but constant advantage in performance. OD is important for the algorithm. Deactivating it makes training more noisy and unstable and gives worse results. Deactivating VR also makes training unstable.}
\label{fig:one_component}
\end{figure}

In Figure~\ref{fig:one_component}, we disable each of the four components of the algorithm. We
find that BN provides a small, but constant advantage in performance. OD is also important for the algorithm.
Disabling OD makes training more noisy and unstable and gives worse results. Disabling VR also
makes training unstable. AG has the least effect in the performance of the algorithm. Furthermore,
disabling more than one component makes training even more unstable in the majority of scenarios.
A summary of the results is available in Table~\ref{tab:results}. In all cases, we use early stopping on the validation log-loss. Furthermore, we present the train log-loss corresponding to the best validation loss as well. Let us note that the log-loss is computed per data point.

\subsection{PTB Character-level LM}

\begin{figure}[htp]
\centering
\includegraphics[width=10cm]{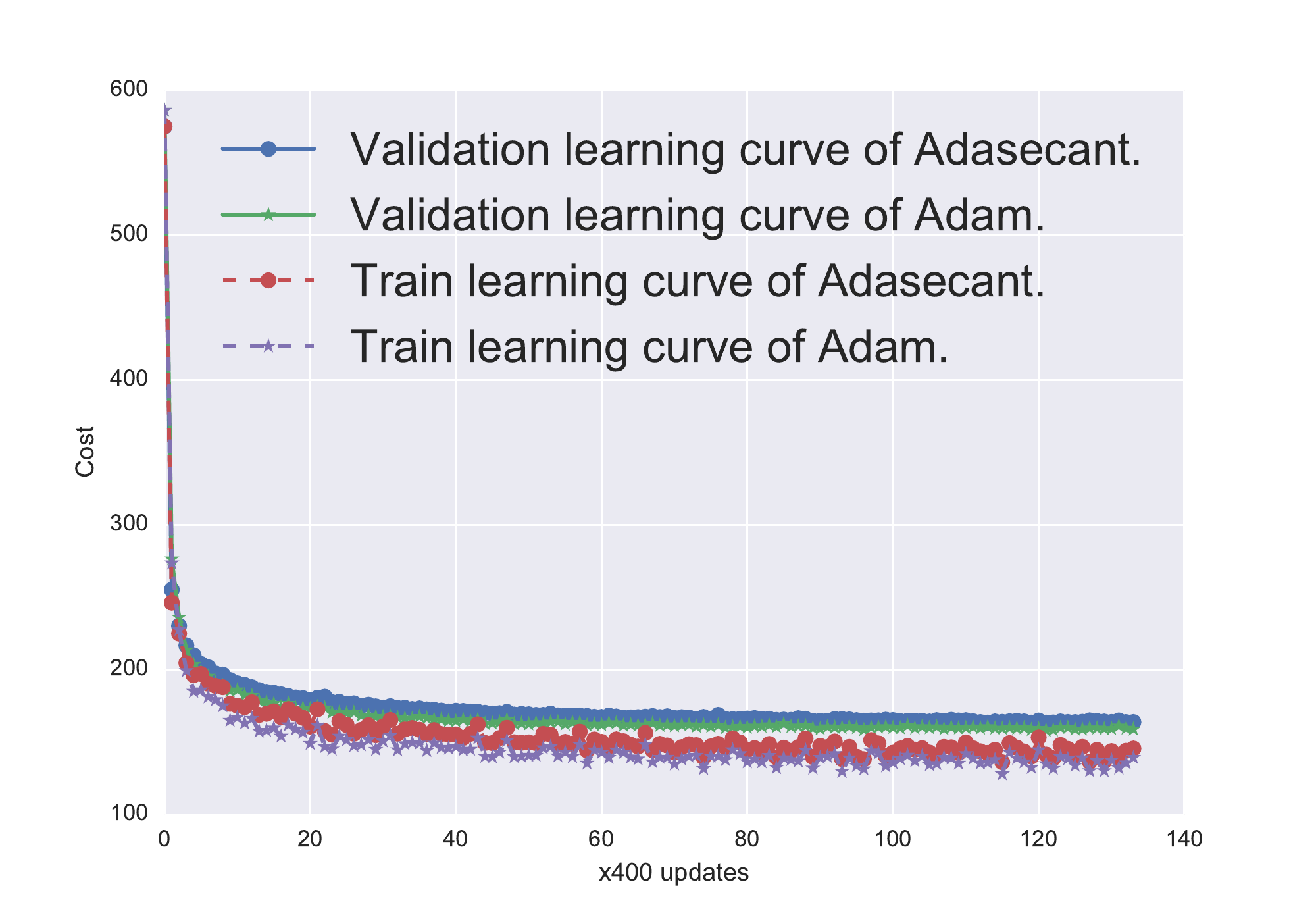}
\caption{Learning curves for the very well-tuned Adam vs AdaSecant algorithm without any
hyperparameter tuning. AdaSecant performs very close to the very well-tuned Adam on PTB
character-level language modeling task. This shows us the robustness of the algorithm to its
hyperparameters.}
\label{fig:ptb_adas_adam}
\end{figure}

We have run experiments with GRU-RNN\cite{cho2014learning} on PTB dataset for character-level language
modeling over the subset defined in \cite{mikolov2012subword}. On this task, we use 400
GRU units with minibatch size of $20$. We train the model over the sequences of length $150$.
For AdaSecant, we have not run any hyperparmeter search, but for Adam we run a hyperparameter search for the
learning rate and gradient clipping. The learning rates are sampled from log-uniform distribution
between $1e-1$ and $6e-5$. Gradient clipping threshold is sampled uniformly between $1.2$ to $20$.
We have evaluated $20$ different pairs of randomly-sampled learning rates and gradient clipping thresholds.
The rest of the hyper-parameters are fixed to their default values. We use the model with the best
validation error for Adam. For AdaSecant algorithm, we fix all the hyperparameters to their default values.
The learning curves for the both algorithms are shown in Figure \ref{fig:ptb_adas_adam}.

\subsection{MNIST with Maxout Networks}
\begin{figure}[ht!]
\centering
\subfigure[$2$ layer Maxout Network]{
\label{fig:depthexpa}
\includegraphics[width=8cm]{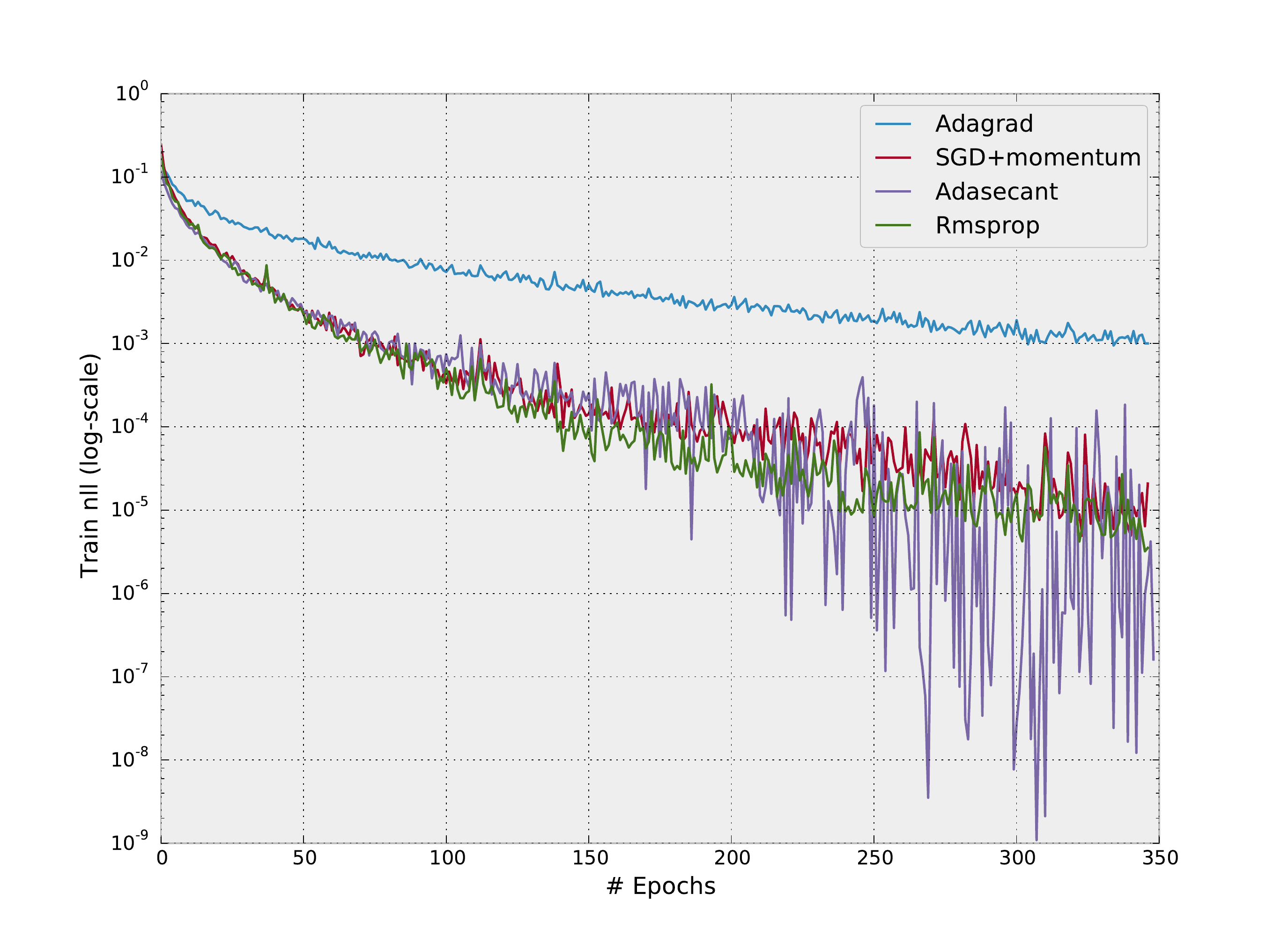}
}
\subfigure[$16$ layer Maxout Network]{
\label{fig:depthexpb}
\includegraphics[width=8cm]{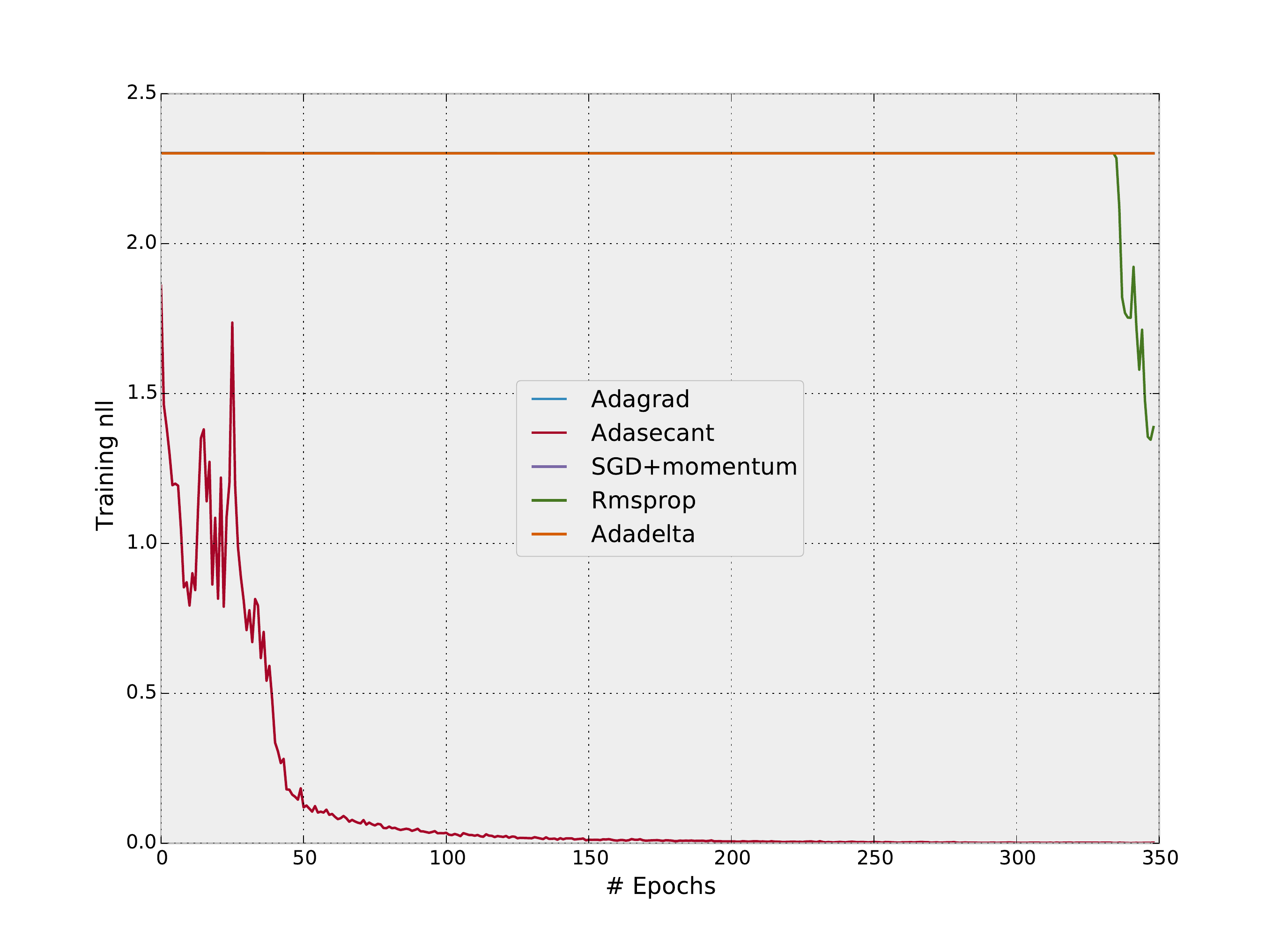}
}
\caption{Comparison of different stochastic gradient algorithms on MNIST with Maxout Networks.
    Both a) and b) are trained with dropout and maximum column norm constraint regularization on
    the weights. Networks are initialized with weights sampled from a Gaussian distribution with
    $0$ mean and standard deviation of $0.05$. In both experiments, the proposed algorithm,
    AdaSecant, seems to be converging faster and arrives to a better minima in training set. We trained both
    networks for $350$ epochs over the training set.}
\label{fig:depthexp}
\end{figure}

The results are summarized in Figure~\ref{fig:depthexp} and we show that AdaSecant converges
as fast or faster than other techniques, including the use of hand-tuned global learning rate and momentum for SGD,
RMSprop, and Adagrad. In our experiments with AdaSecant algorithm, adaptive momentum term $\gamma_i^k$ was clipped at
$1.8$. In $2$-layer Maxout network experiments for SGD-momentum experiments, we used the best hyper-parameters reported
by \cite{goodfellow2013maxout}, for RMSProp and Adagrad, we crossvalidated learning rate for $15$ different learning rates
sampled uniformly from the log-space. We crossvalidated $30$ different pairs of momentum and learning rate for SGD+momentum,
for RMSProp and Adagrad, we crossvalidated $15$ different learning rates sampled them from log-space uniformly
 for deep maxout experiments.

\section{Conclusion}
We described a new stochastic gradient algorithm with adaptive learning rates
that is fairly insensitive to the tuning of the hyper-parameters and doesn't require tuning of learning
rates. Furthermore, the variance reduction technique we proposed improves the
convergence when the stochastic gradients have high variance.
Our algorithm performs as well or better than other popular, carefully-tuned stochastic gradient algorithms. We also present a comprehensive ablation study where we show the effects and importance of each of the elements of our algorithm. As future work, we should try to find theoretical convergence properties  of the algorithm to understand it better analytically.

\section*{Acknowledgments}

We thank the developers of Theano \cite{Bastien-2012}, Pylearn2 \cite{pylearn2_arxiv_2013} and Blocks \cite{Merrienboer2015} and the computational resources provided by Compute Canada and Calcul Qu\'ebec. This work has been partially supported by NSERC, CIFAR, and Canada Research Chairs, Project TIN2013-41751, grant 2014-SGR-221. Jose Sotelo also thanks the Consejo Nacional de Ciencia y Tecnolog\'ia (CONACyT) as well as the Secretar\'ia de Educaci\'on P\'ublica (SEP) for their support. We would like to thank Tom Schaul for the valuable discussions. We also thank Kyunghyun Cho and Orhan Firat for proof-reading and giving feedbacks on the paper.

\nocite{theano2016}
\nocite{pylearn2_arxiv_2013}
\nocite{Merrienboer2015}

\bibliographystyle{IEEEtran}
\bibliography{submission_nips_w}

\appendix
\section{Appendix}
\subsection{Derivation of Equation~\ref{eqn:exp_gamma_cri}}
\begin{align*}
\label{eqn:exp_gamma_opt2}
\frac{\partial \E[(\beta_i g_i + (1 - \beta_i) \E[g_i] - g_i^{\prime})^2]}{\partial \beta_i} + \lambda \beta_i^2&=0
\end{align*}
\begin{align*}
\E[&(\beta_i g_i + (1 - \beta_i) \E[g_i] - g_i^{\prime}) \\
&\frac{\partial (\beta_i g_i + (1 - \beta_i) \E[g_i] - g_i^{\prime})}{\partial \beta_i}] + \lambda \beta_i=0
\end{align*}
\begin{align*}
\E[(\beta_i g_i + (1 - \beta_i) \E[g_i] - g_i^{\prime})(g_i -  \E[g_i])] + \lambda \beta_i=0
\end{align*}
\begin{align*}
\E[&(\beta_i g_i (g_i -  \E[g_i]) + (1 - \beta_i) \E[g_i] (g_i - \E[g_i]) & \\
 &- g_i^{\prime}(g_i - \E[g_i])] + \lambda \beta_i&=0 \\
\end{align*}
\begin{align*}
\beta_i & = \frac{\E[(g_i - \E[g_i])(g_i^{\prime} - \E[g_i])]}{\E[(g_i - \E[g_i])(g_i - \E[g_i])] + \lambda} \\
& = \frac{\E[(g_i - \E[g_i])(g_i^{\prime} - \E[g_i])]}{\var(g_i) + \lambda}
\end{align*}

\subsection{Further Experimental Details}
\begin{figure}[ht!]
\centering
\includegraphics[width=10cm]{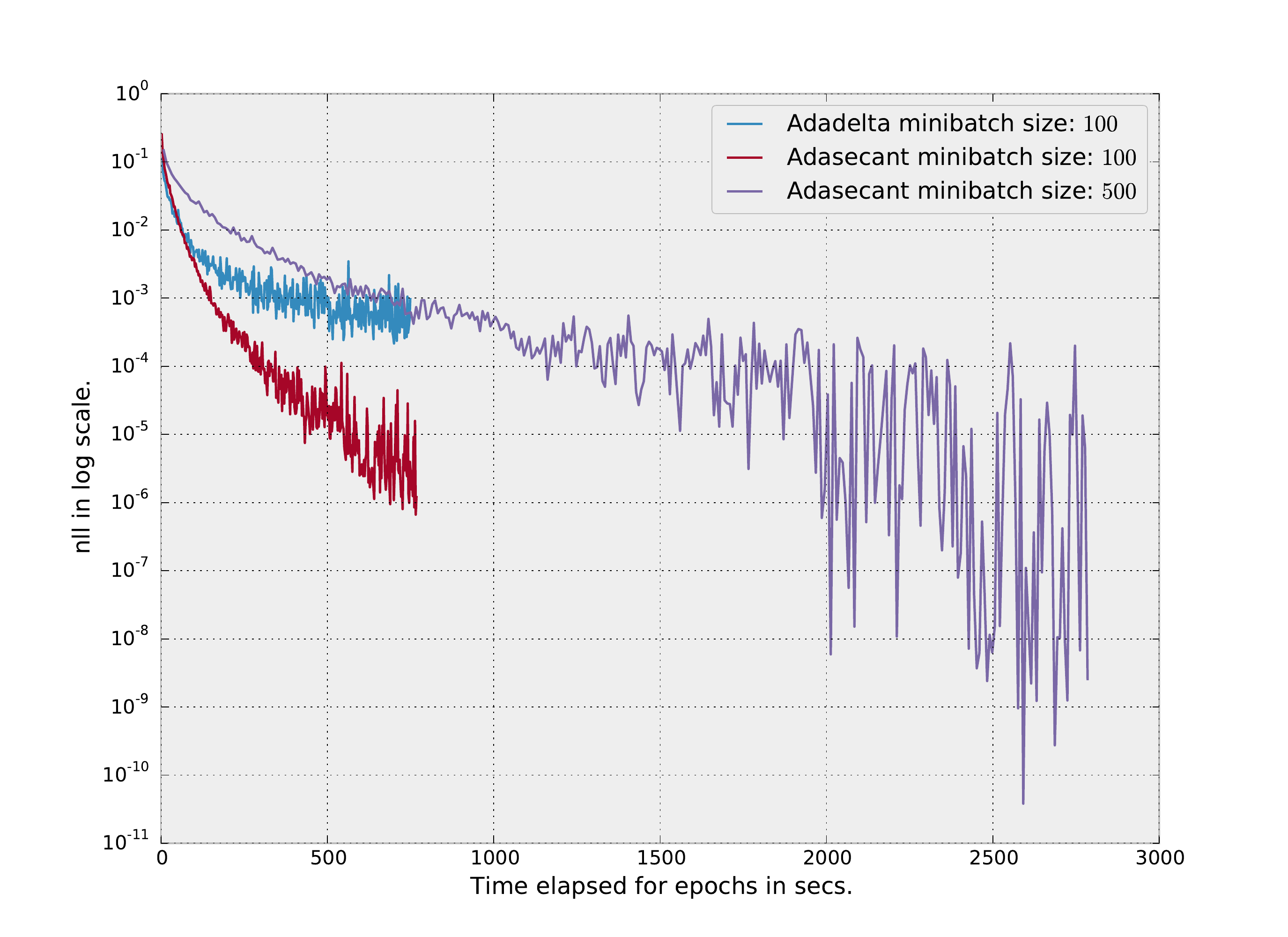}
\caption{In this plot, we compared AdaSecant trained by using minibatch size of $100$ and $500$
    with adadelta using minibatches of size $100$. We performed these experiments on MNIST with
    2-layer maxout MLP using dropout.}
\label{fig:adasecant_clock}
\end{figure}

In Figure~\ref{fig:adasecant_clock}, we analyzed the effect of using
different minibatch sizes for AdaSecant and compared its convergence with Adadelta in wall-clock
time. For minibatch size $100$ AdaSecant was able to reach the almost same training negative log-likelihood
as Adadelta
after the same amount of time, but its convergence took much longer. With minibatches of size $500$ AdaSecant
was able to converge faster in wallclock time to a better local minima.

\subsection{More decomposition experiments}
\begin{figure}[h]
\centering
\includegraphics[width=6cm]{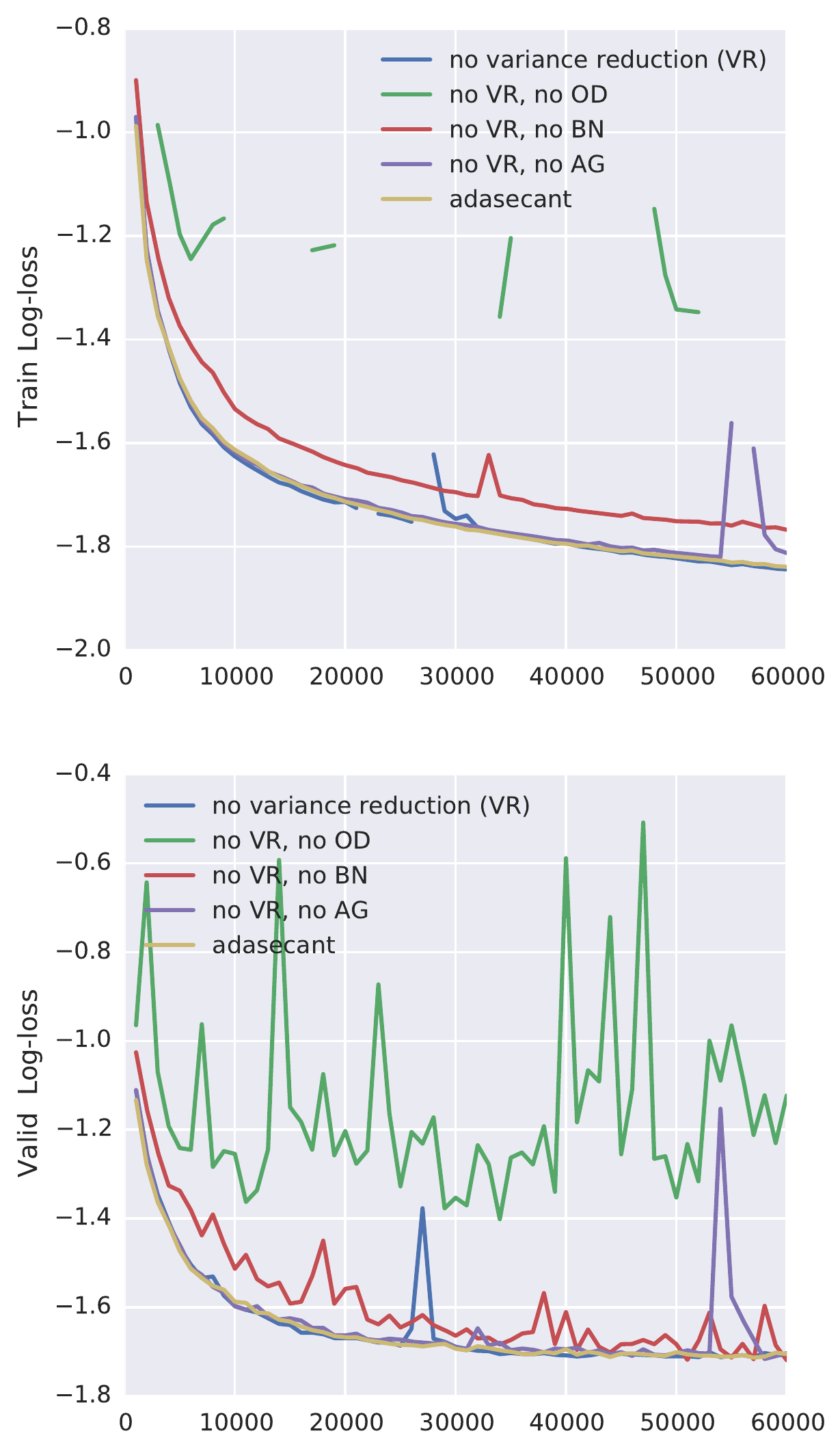}
\caption{No variance reduction comparison.}
\label{fig:variance_reduction}
\end{figure}

\begin{figure}[h]
\centering
\includegraphics[width=6cm]{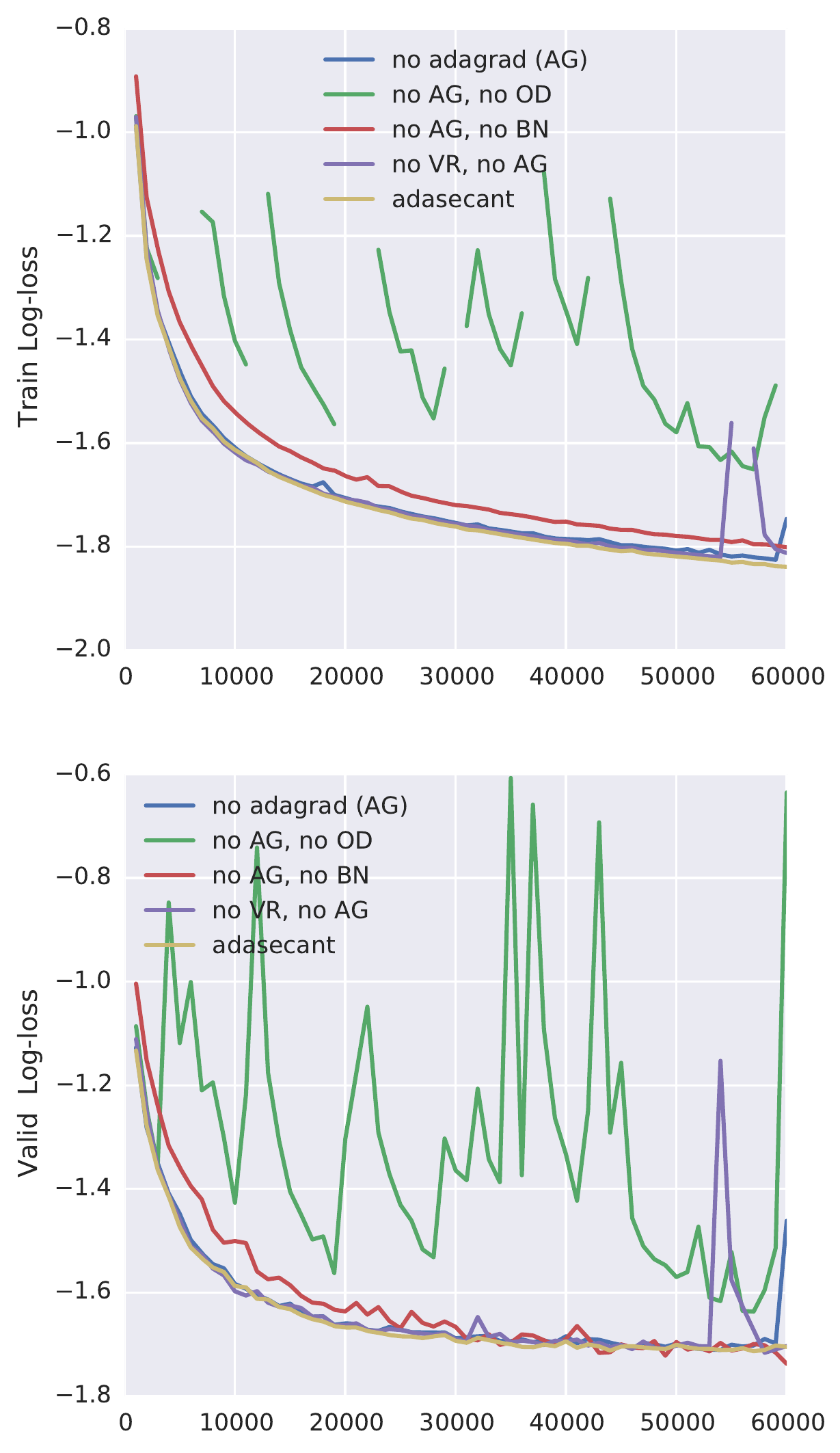}
\caption{No Adagrad comparison.}
\label{fig:adagrad}
\end{figure}

\begin{figure}[ht!]
\centering
\includegraphics[width=6cm]{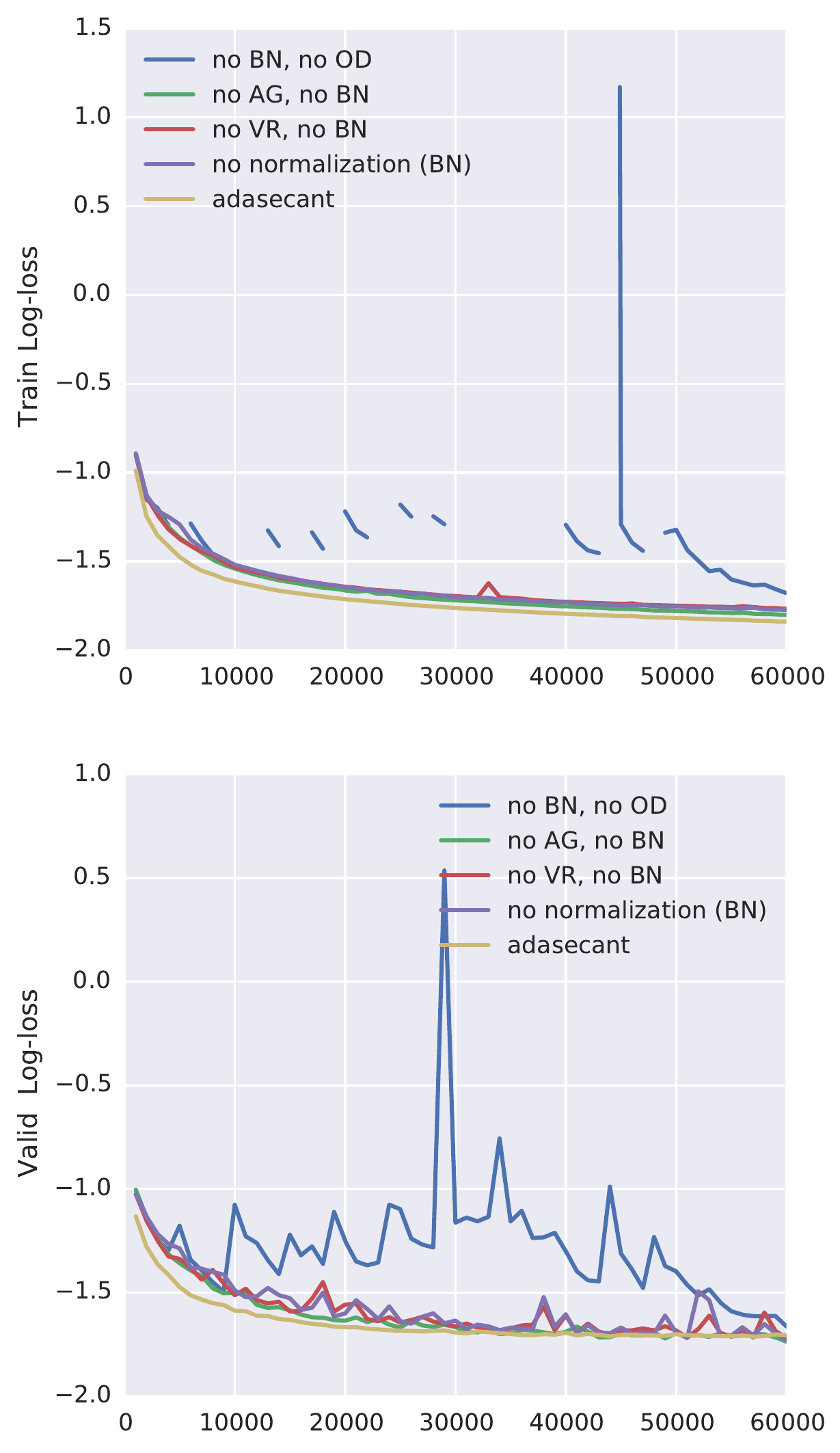}
\caption{No block normalization comparison.}
\label{fig:block_normalization}
\end{figure}

\begin{figure}[h]
\centering
\includegraphics[width=6cm]{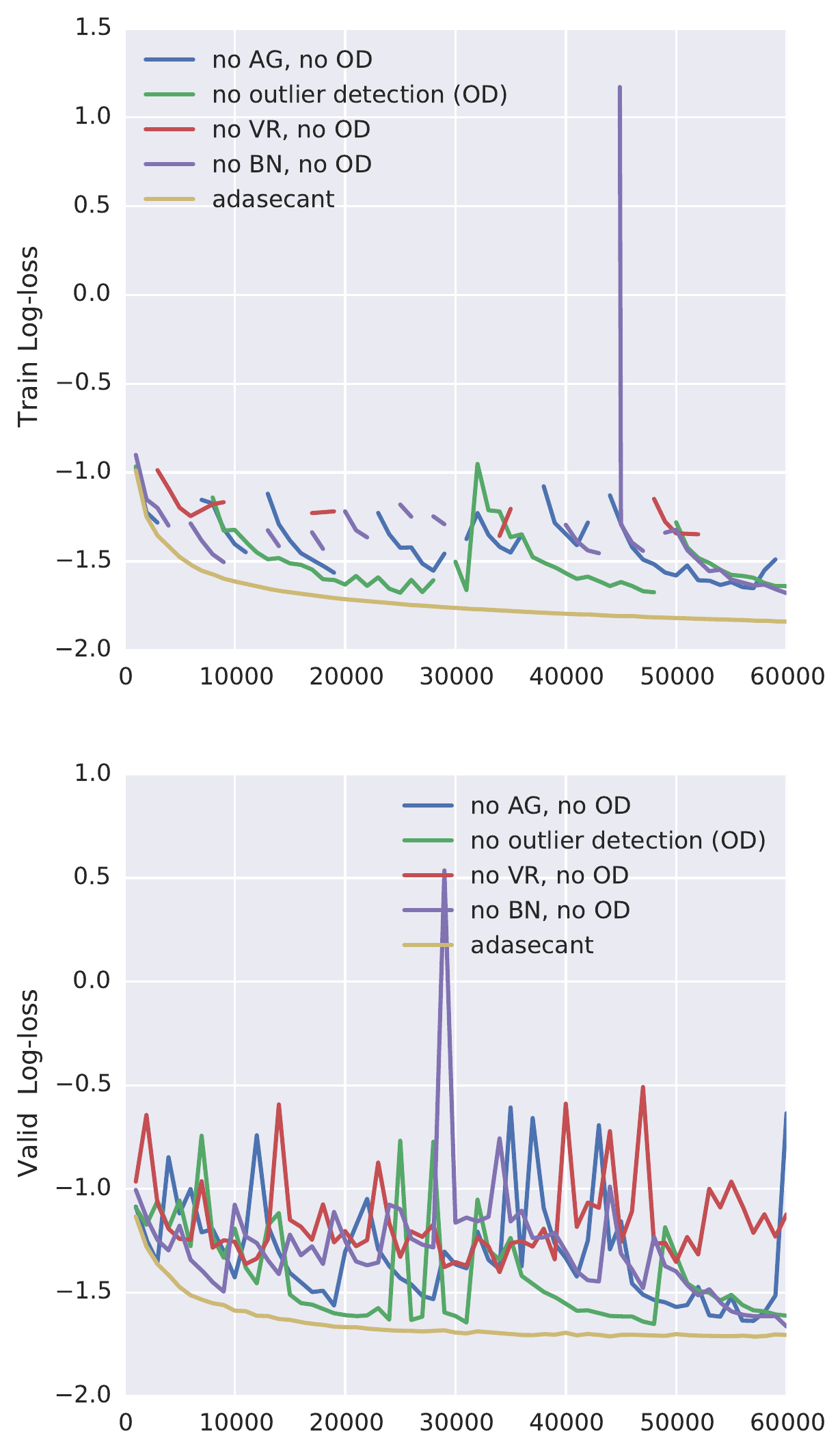}
\caption{No outlier detection comparison.}
\label{fig:outlier_detection}
\end{figure}

We have run experiments with the different combinations of the components of the algorithm. We
show those results on handwriting synthesis with IAM-OnDB dataset. The results can be observed from
Figure~\ref{fig:variance_reduction}, Figure~\ref{fig:adagrad},
Figure~\ref{fig:block_normalization}, and Figure~\ref{fig:outlier_detection} deactivating the components leads to a more unstable training
curve in the majority of scenarios.

\end{document}